\documentclass[conference]{IEEEtran}
\IEEEoverridecommandlockouts

\usepackage{amsmath,amssymb,amsfonts}
\usepackage{mathtools}
\usepackage{bm}
\usepackage{graphicx}
\usepackage{xcolor}
\usepackage{cite}
\usepackage{algorithmic}
\usepackage{array}
\usepackage{url}
\usepackage{amsthm}
\usepackage{hyperref}
\usepackage{comment}
\usepackage{wrapfig}
\hypersetup{colorlinks=true, linkcolor=blue, citecolor=blue, urlcolor=blue}

\newcommand{\cW}{\mathcal{W}}
\newcommand{\cX}{\mathcal{X}}
\newcommand{\cG}{\mathcal{G}}
\newcommand{\cV}{\mathcal{V}}
\newcommand{\cE}{\mathcal{E}}
\newcommand{\bW}{\mathbf{W}}
\newcommand{\bX}{\mathbf{X}}
\newcommand{\bZ}{\mathbf{Z}}
\newcommand{\bI}{\mathbf{I}}
\newcommand{\hbW}{\hat{\mathbf{W}}}

\title{A Framework for Directed Acyclic Hypergraph Learning
\thanks{
Z. Dong is with the Institute for Financial Services Analytics,
University of Delaware, Newark, DE 19716, USA.
C. Mundo-Levano and G. R. Arce are with the Department of Electrical and Computer Engineering,
University of Delaware, Newark, DE 19716, USA.
W. Qian with the Department of Applied Economics AND Statistics,
University of Delaware, Newark, DE 19716, USA.
D. Lau is with the Department of Electrical and Computer Engineering,
University of Kentucky, Lexington, KY 40506, USA.
This work was partially supported by the National Science Foundation under grants 2230161, 1815992, and 1816003, and by the Institute of Financial Services Analytics, co-sponsored by JP Morgan Chase \& Co.}
}
\author{
  \IEEEauthorblockN{Zhiyuan Dong, Carlos Mundo-Levano, Wei Qian, Daniel Lau, Gonzalo R.\ Arce}
}

\begin{document}

\maketitle

\begin{abstract}
Continuous optimization methods for learning Directed Acyclic Graphs (DAGs)
operate on weighted adjacency matrices and are
therefore limited to pairwise causal relationships. We propose a framework for
learning Directed Acyclic Hypergraphs (DAHGs) from observational data,
capturing joint parental influences that pairwise models cannot represent. Our
approach rests on three components: (i) a generalized linear structural
equation model (SEM) with multiplicative interaction terms whose non-zero
weights correspond one-to-one with directed hyperedges; (ii) a weighted
adjacency tensor representation whose acyclicity is characterized via
nilpotency under the tensor t-product; and (iii) a differentiable acyclicity
constraint derived through the Fourier decomposition of the t-product, which
reduces tensor nilpotency to slice-wise matrix nilpotency and enables
least-squares learning via the augmented Lagrangian method.
\end{abstract}

\section{Introduction}

Learning causal structure from observational data is a central problem in
statistical machine learning. Continuous optimization approaches
\cite{zheng2018dags, yu2019dag, bello2022dagma, ng2022masked, lee2019scaling, rey2025non} have made
DAG structure learning tractable by reformulating the combinatorial acyclicity
constraint as a smooth, differentiable function of the weighted adjacency
matrix~$\bW$. Under the linear SEM $\bX = \bW^\top \bX + \bZ$, the structure
learning problem becomes a least-squares program subject to the acyclicity
constraint $h(\bW) = 0$.

These methods are inherently limited to \emph{pairwise} causal relationships.
In many domains such as gene regulation~\cite{klamt2009hypergraphs}, neural
circuits~\cite{yu2011higher}, and economic networks~\cite{acemoglu2012network}, a target variable
depends on the \emph{joint} action of multiple parents, an interaction that
cannot be decomposed into a sum of individual effects. The natural graphical
object for such interactions is a \emph{directed hypergraph}, in which each
hyperedge has a single head node and a set of tail nodes whose joint influence
it encodes. Extending continuous DAG learning to directed acyclic hypergraphs
(DAHGs) requires: (a) a structural equation model that captures
multiplicative interactions, (b) a higher-order representation of the
adjacency structure, and (c) a differentiable acyclicity constraint that
operates on this representation.

While recent work~\cite{enouen2025higher} has established identifiability of DAHG structures
under additive noise, no continuous optimization framework exists.  We develop a tensor-based framework that addresses all
three requirements. The key technical insight is that the tensor
t-product~\cite{kilmer2011factorization} provides a natural algebraic setting
in which adjacency, acyclicity, and the SEM can be jointly formulated, and in
which the Fourier decomposition of the t-product reduces the tensor acyclicity
condition to independent matrix conditions on the Fourier slices, enabling
direct generalization of existing DAG constraints such as DAGMA~\cite{bello2022dagma}.

\section{Generalized SEM and Hypergraph Structure}

\subsection{From Linear SEM to Joint Interactions}

The standard linear SEM models each node as a weighted sum of its parents:
$X_i = \sum_{j \in \mathrm{Pa}(i)} W_{ij} X_j + Z_i$. To capture joint
parental influences, we introduce multiplicative interaction terms. Let
$P_i^{(r)}$ be the tail set of the $r$-th hyperedge with head $i$. The
generalized SEM is
\begin{equation}
X_i = \sum_{r=1}^{K_i} \cW_{i,\, P_i^{(r)}} \prod_{j \in P_i^{(r)}} X_j
\;+\; Z_i,
\label{eq:gen_sem}
\end{equation}
where $K_i$ is the number of hyperedges with head $i$ and
$\cW_{i,\, P_i^{(r)}}$ encodes the strength of the joint influence
of $P_i^{(r)}$ on node $i$. When every $|P_i^{(r)}| = 1$,
\eqref{eq:gen_sem} reduces to the standard linear SEM.

\subsection{Hypergraph Interpretation and Tensor Representation}

Each non-zero weight $\cW_{i,\, P_i^{(r)}}$ in~\eqref{eq:gen_sem}
corresponds to a directed hyperedge $(P_i^{(r)} \to i)$ in a DAHG
$\cG = (\cV, \cE)$. Following~\cite{DAHG2026tsvd}, we represent the hypergraph as a weighted
adjacency tensor $\cW \in \mathbb{R}^{N \times N \times \cdots \times N}$
of order~$M$, where the first mode indexes head nodes and modes $2$
through~$M$ jointly index tail configurations. For each hyperedge $e$
with head $v_i$, tail set $S_e$, and weight $\omega_e \in \mathbb{R}$,
\begin{equation}
\cW_{i,\, p_2, \ldots, p_M} = \frac{(c_e - 1)\;\omega_e}{\alpha_e},
\label{eq:tensor_entry}
\end{equation}
where $c_e = |S_e| + 1$ and $\alpha_e$ is the multinomial normalization
over tail-index permutations~\cite{DAHG2026tsvd}. When
$M = 2$, $\cW$ reduces to the standard weighted adjacency matrix.
\begin{figure}[t]
\centering
\includegraphics[width=\columnwidth]{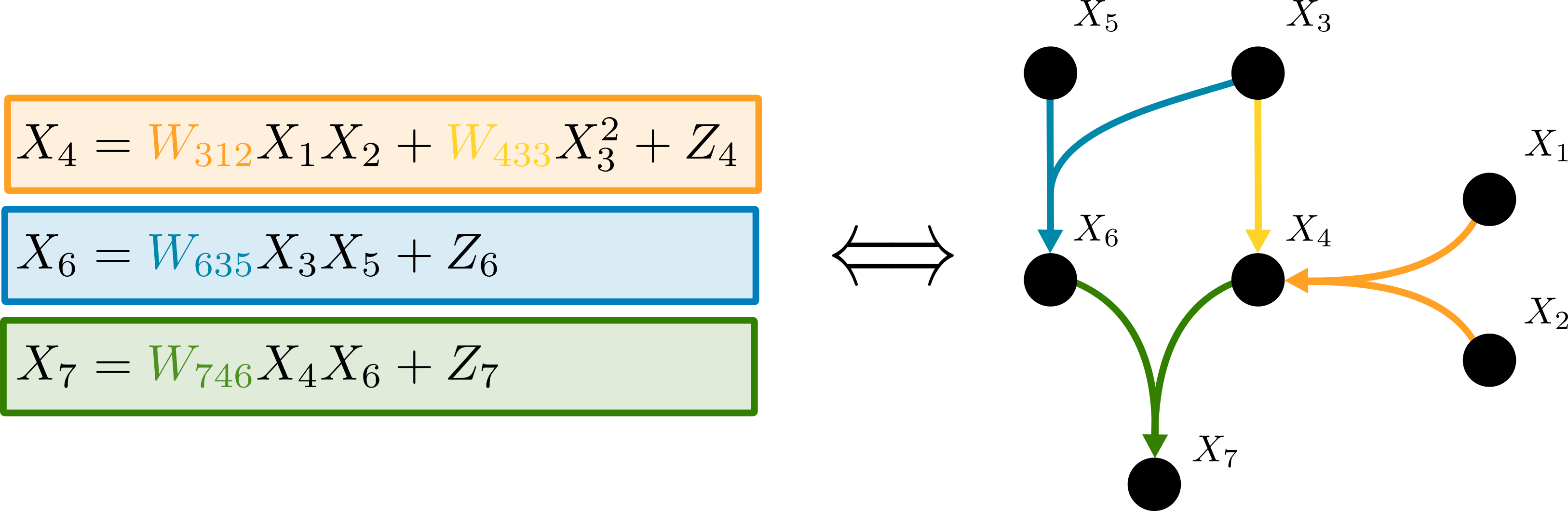}
\caption{Illustration of the generalized SEM and its DAHG interpretation.
\textbf{Left:} Three structural equations with multiplicative interaction
terms. Node $X_4$ is influenced by two hyperedges: $\{X_1, X_2\} \to X_4$
with weight $W_{312}$ and $X_3 \to X_4$ with
weight $W_{433}$. Nodes $X_6$ and $X_7$ each have a single order-3
hyperedge. \textbf{Right:} The corresponding DAHG, where each colored
hyperedge connects a tail set (source nodes) to a single head node.}
\label{fig:sem_dahg}
\end{figure}
Using the t-product~\cite{kilmer2011factorization}, the generalized SEM
across all $N$ nodes and $D$ feature dimensions can be written compactly as
\begin{equation}
\bX = \bigl(\cW \ast \cX\bigr)^{(1)} + \bZ,
\label{eq:tensor_sem}
\end{equation}
where $\cX \in \mathbb{R}^{N \times D \times N^{M-2}}$ is the cross-node
interaction tensor defined in~\cite{wang2024t}, $\ast$ denotes the
t-product, $(\cdot)^{(1)}$ extracts the first frontal slice, and
$\mathbf{Z} \in \mathbb{R}^{N \times D}$ is the noise matrix.
When $M = 2$, $\cW$ reduces to the standard adjacency matrix and
\eqref{eq:tensor_sem} reduces to the standard linear SEM.

\section{Differentiable Acyclicity Constraint and Learning}

Under the t-product, the $k$-th power $\cW^k = \cW \ast \cW \ast \cdots
\ast \cW$ acts as matrix multiplication on the first two modes (head and
tail) independently across each channel (modes $3$ through~$M$). The
$(i,j)$-th slice of $\cW^k$ aggregated across channels encodes directed
walks of length~$k$ from $j$ to $i$ in the clique-expanded representation
of the hypergraph. The DAHG is acyclic if and only if $\cW$ is nilpotent
under the t-product: $\cW^k = \mathbf{0}$ for some $k \leq N$.

Directly checking tensor nilpotency is intractable. The key simplification
comes from the Fourier decomposition of the t-product: letting
$\hat{\cW} = \mathrm{fft}(\cW, [], 3)$ with frontal slices $\hbW^{(j)}$
for $j = 1, \ldots, n_s$, the t-product decomposes into independent matrix
multiplications across Fourier slices. The tensor $\cW$ is nilpotent under
the t-product if and only if every Fourier slice $\hbW^{(j)}$ is nilpotent
as a matrix. In the general case, nilpotency must be enforced on each slice
independently. Since $\hbW^{(j)}$ is generally complex for $j \geq 2$, we
replace the standard Hadamard square with the element-wise modulus square
$\hbW^{(j)} \circ \bar{\hbW}^{(j)}$, whose $(a,b)$-entry is
$|(\hbW^{(j)})_{ab}|^2 \geq 0$. This matrix is real and non-negative,
shares the same sparsity pattern as $\hbW^{(j)}$, and is nilpotent if and
only if $\hbW^{(j)}$ is nilpotent. Applying the DAGMA log-determinant
characterization~\cite{bello2022dagma} to each slice yields
\begin{equation}
h_{\mathrm{full}}(s, \cW) = \sum_{j=1}^{n_s} \left[ n \log s -
\log\det\!\left(s\bI - \hbW^{(j)} \circ \bar{\hbW}^{(j)}\right) \right]
= 0,
\label{eq:logdet_full}
\end{equation}
which is differentiable with respect to $\cW$ since $\hbW^{(j)}$ is a
linear function of the real tensor $\cW$ via the FFT. This constraint is
valid without any sign assumption on $\cW$, but requires $n_s = N^{M-2}$
log-determinant evaluations. When all tensor entries are non-negative
($\cW \geq 0$), a natural condition since hyperedge weights $\omega_e > 0$,
a significant simplification is possible. The first Fourier slice
$\hbW^{(1)} = \sum_{i=1}^{n_s} \bW^{(i)}$ is real and non-negative. If
$\hbW^{(1)}$ is nilpotent, then by non-negativity every mixed product
$\bW^{(i_1)} \cdots \bW^{(i_k)} = \mathbf{0}$, which forces all remaining
Fourier slices to be nilpotent as well. The full tensor nilpotency condition
thus reduces to nilpotency of a single real matrix, and the constraint
simplifies to
\begin{equation}
h(s,\, \cW) = n \log s - \log\det\!\left(s\bI - \hbW^{(1)}\right) = 0,
\label{eq:logdet}
\end{equation}
where $s > \rho(\hbW^{(1)})$. This reduction yields two practical
advantages: the computational cost drops from $n_s = N^{M-2}$ complex
matrix log-determinants to a single real one, and the resulting subproblem
is convex in $\hbW^{(1)}$~\cite{rey2025non}.

With the acyclicity constraint in hand, we formulate the DAHG structure
learning problem. Given $T$ i.i.d.\ observations $\bX \in
\mathbb{R}^{N \times T}$, the DAHG
structure learning problem is formulated as
\begin{equation}
\min_{\cW \geq 0} \;\; \frac{1}{2T} \bigl\|\bX - (\cW \ast
\cX)^{(1)}\bigr\|_F^2 + \lambda \|\cW\|_1
\quad \text{s.t.} \quad h(s,\, \cW) = 0,
\label{eq:objective}
\end{equation}
where $h(s, \cW)$ is defined in~\eqref{eq:logdet}, the $\ell_1$ penalty
promotes sparsity of the learned hypergraph, and the non-negativity constraint
$\cW \geq 0$ ensures that the Fourier-domain reduction to a single real
matrix is valid. The constraint is enforced via the augmented Lagrangian
method~\cite{bonnans2006numerical}. When $M = 2$, the optimization problem reduces exactly to DAG Learning Problem.

\section{Experiments}

We validate both constraints on synthetic ladder-style DAHGs with genuine
order-3 hyperedges and random weights greater than 1.
Figure~\ref{fig:val}(a) shows that both $h$ and $h_{\mathrm{full}}$ are
strictly positive for cyclic configurations and zero for acyclic ones,
with $h_{\mathrm{full}}$ decaying more slowly for long cycles.
Figure~\ref{fig:val}(b) adds random negative-weight hyperedges to a fixed
cyclic hypergraph: $h$ drops below zero within a few additions, while
$h_{\mathrm{full}}$ remains positive throughout, confirming the necessity
of $\cW \geq 0$ for the single-matrix reduction. We also apply the framework to the protein-signaling
dataset~\cite{sachs2005causal}, using a refined ground
truth~\cite{kleinegesse2022domain}
(Figure~\ref{fig:protein}). The method matches 13 of 17 ground-truth
edges and recovers biologically plausible hyperedges such as
$\{$PKA, Mek$\} \to$ Raf and $\{$PIP2,
Plc$\gamma\} \to$ PIP3, consistent with known signaling
interactions~\cite{balan2006identification,rhee2001regulation, tariq2021striking}, demonstrating the
value of modeling joint causal mechanisms beyond pairwise edges.

\section{Conclusion}

We presented a tensor-based framework for learning DAHGs from
observational data, where the Fourier decomposition of the t-product
yields differentiable acyclicity constraints that generalize existing
DAG constraints. Under non-negative weights, the constraint reduces to
a single convex subproblem. Future work includes variational extensions
for latent confounders and low-rank approximations for higher hyperedge
orders.

\bibliographystyle{IEEEtran}
\bibliography{bibliography}

\newpage
\appendix
\section*{Appendix: Figures and Tables}

\begin{figure}[h]
\centering
\includegraphics[width=0.49\columnwidth]{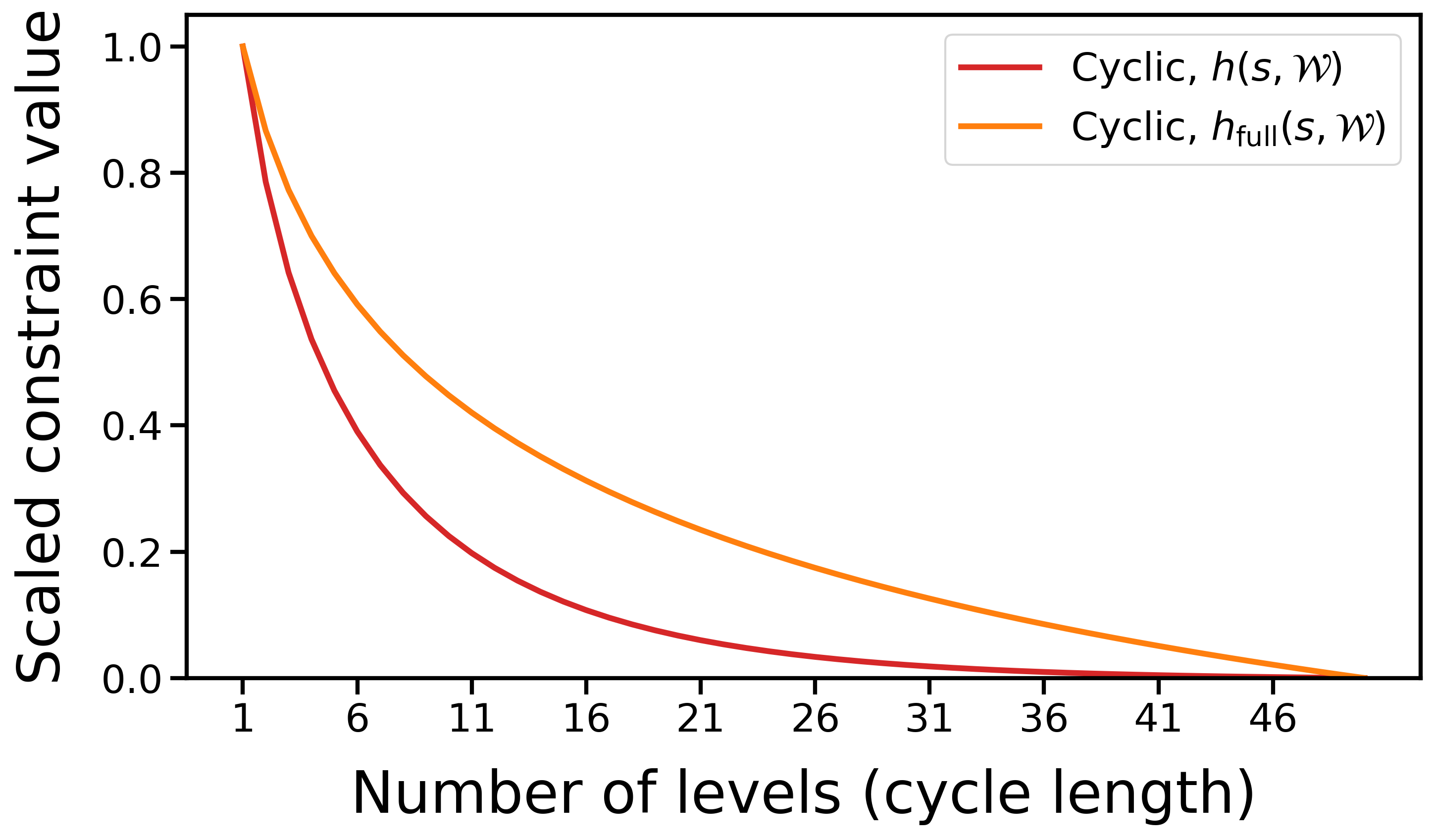}
\hfill
\includegraphics[width=0.49\columnwidth]{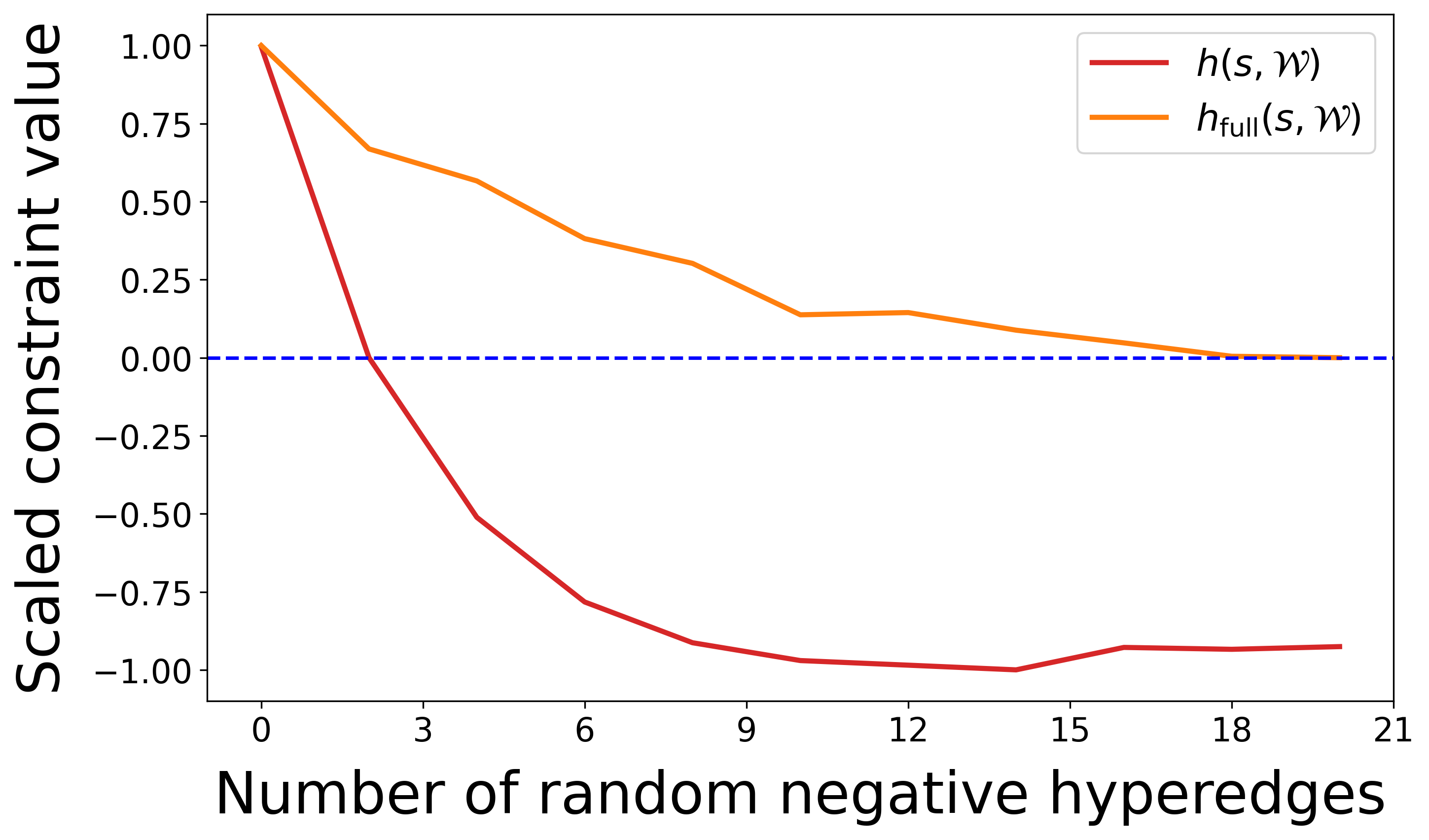}
\caption{Validation on ladder DAHGs with order-3 hyperedges and random
weights greater than 1. \textbf{Left:} Both constraints are positive for
cyclic and zero for acyclic (not shown); $h_{\mathrm{full}}$ decays more
slowly as it penalizes high-weight cycles more strongly. \textbf{Right:}
Negative hyperedges added to a length-10 cycle. $h$ drops below zero;
$h_{\mathrm{full}}$ remains positive, confirming the necessity of
$\cW \geq 0$ for the single-matrix reduction.}
\label{fig:val}
\end{figure}

\begin{figure}[h]
\centering
\includegraphics[width=0.75\columnwidth]{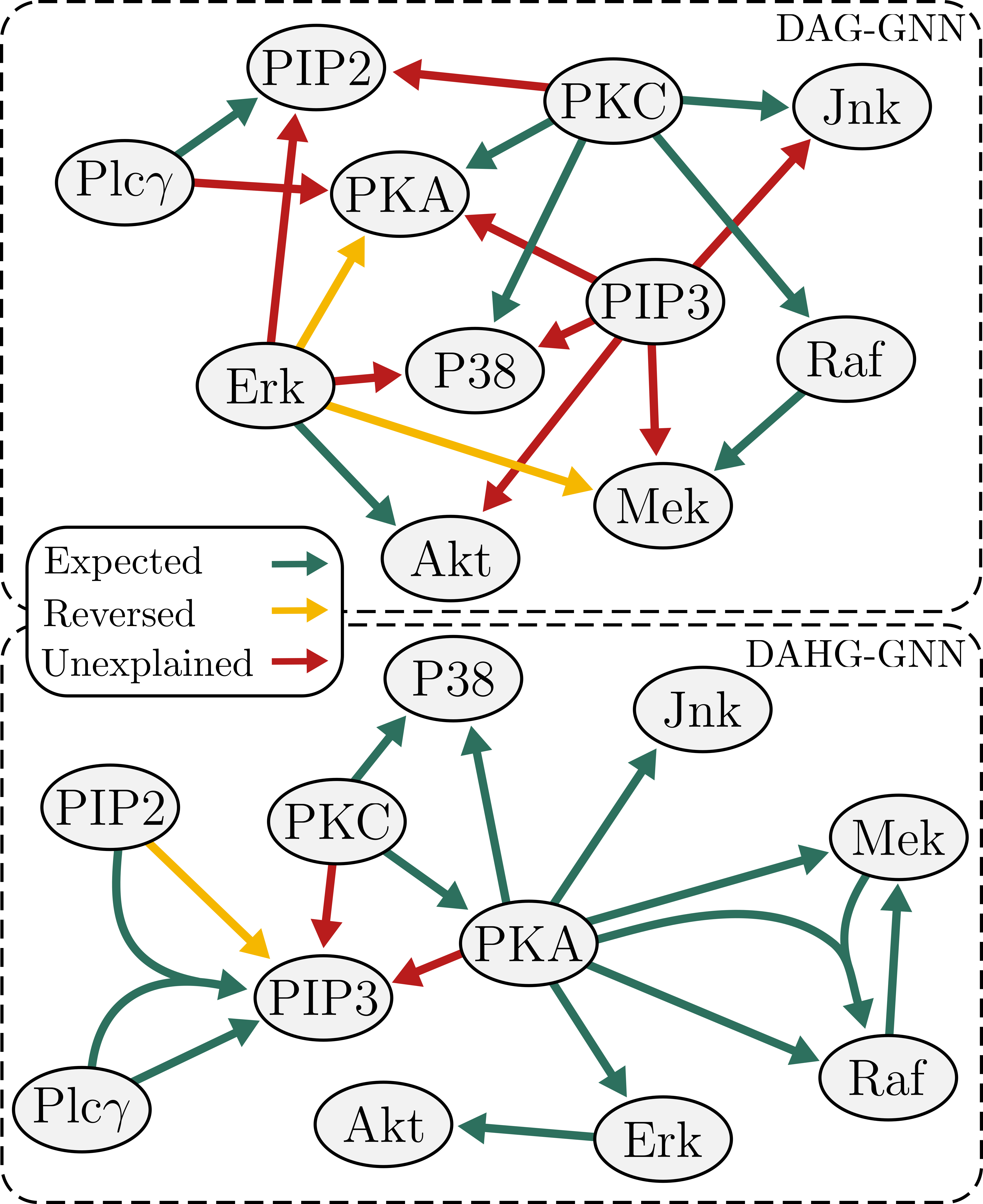}
\caption{Comparison of learned protein directed (hyper)graphs. Top: DAGGNN~\cite{yu2019dag} Bottom: our DAHG method. Connections are colored by accuracy:
green (expected), yellow (reversed), and red (unexplained). Our method
matches 13 of 17 refined ground-truth edges\cite{kleinegesse2022domain} and recovers hyperedges that align
with known multi-protein signaling structures.}
\label{fig:protein}
\end{figure}

\end{document}